\documentclass[runningheads]{llncs}
\usepackage{graphicx}
\usepackage{multirow}
\begin{document}
\title{Using Issues to Explain Legal Decisions}
%
%
\author{Trevor Bench-Capon}
\authorrunning{Trevor Bench-Capon}
%
\institute{University of Liverpool, Liverpool, UK.
\email{tbc@csc.liv.ac.uk}\\
}
\maketitle              
\begin{abstract}
The need to explain the output from Machine Learning systems designed to predict the outcomes of legal cases has led to a renewed interest in the explanations offered by traditional AI and Law systems, especially those using factor based reasoning and precedent cases. In this paper we consider what sort of explanations we should expect from such systems, with a particular focus on the structure that can be provided by the use of issues in cases.

\end{abstract}
\section{Introduction}

Currently there is much interest in the use of Machine Learning (ML) based approaches to predict legal decisions. The European Convention on Human Rights alone has been the subject of a cluster of such systems including \cite{aletras2016predicting}, \cite{medvedeva2019using}, \cite{chalkidis2019neural}, \cite{says2020prediction} and \cite{kaur2019convolutional}. Using such systems in legal applications, however, raises a number of issues \cite{bench2020need}, including bias, adapting to changes in statute law, case law and social values, and, perhaps most important, the lack of explanation. In law there is a right to explanation \cite{doshi2017accountability} and so providing explanations that users can understand is essential if AI systems are to be used in practice. This has long been recognised in AI and Law and the provision of explanations has been a central feature of systems developed in the field \cite{atkinson2020explanation}. It is therefore a natural move to see whether it is possible to use the techniques developed to explain the outputs of systems previously developed in AI and Law to explain the outputs of ML systems. In particular the factor based reasoning developed for the CATO system \cite{aleven} and widely adopted in subsequent systems \cite{bench2017hypo} has been proposed as a suitable candidate for this role. In \cite{branting} the idea is to first ascribe factors to cases using ML and then to explain the outcomes in terms of these factors. In \cite{prakken2021ac}, the proposal is to produce the explanation from a case based system running in parallel with the ML system.

In this paper we consider what sort of explanations we can expect from such systems, with a particular focus on the structure that can be provided by the use of issues in cases.

\section{Background: Factor Based Reasoning in CATO}

We begin by describing factor based reasoning in CATO \cite{aleven}, which is the starting point for subsequent accounts of factor based reasoning \cite{bench2017hypo}. CATO is directed towards the domain of US Trade Secrets, and is largely based on the law as set out in the \textit{Restatement of Torts}\footnote{The relevant section, section 757, \textit{Liability for disclosure or use of another's Trade Secret}, can be found at https://www.lrdc.pitt.edu/ashley/restatem.htm.}. In CATO cases are represented as sets of factors. Factors are ascribed on the basis of stereotypical patterns of facts which have legal significance in that they provide a reason to find for one of the parties.  CATO has thirteen factors for each side, as shown in Table 1. The conditions for ascribing them to cases are given in Appendix 2 of \cite{aleven}.
Note that the absence of a factor is not in general a reason to find for the other party: in the rare cases where the absence of a factor might favour the other side, a second distinct factor for the party favoured is used. The only example of this in CATO is \textit{security measures} with factors F6p and F19d. Note, however, that it may be that neither F6p or F19d is present in the case: even if security measures were taken, so that there is no reason to find for the defendant on this aspect, they may not have been sufficient to provide a reason to find for the plaintiff, and so that the aspect is neutral. The factors from \cite{aleven} have been reused by many subsequent researchers, including \cite{bruninghaus2003predicting}, \cite{chorley2005empirical}, \cite{al2016methodology}, \cite{verheij2020} and \cite{prakken2021ac}.

\begin{table}[t]
\caption{Factors in CATO. Numbers are as in \cite{aleven}: ``p'' and ``d'' indicates the side favoured as in \cite{grabmair2017predicting}}
\begin{tabular}{lllll}
\multicolumn{2}{c}{Plaintiff Factors}  &  & \multicolumn{2}{c}{Defendant Factors} \\
F2p  & Bribe-Employee                  &  & F1d   & Disclosure-In-Negotiations    \\
F4p  & Agreed-Not-To-Disclose          &  & F3d   & Employee-Sole-Developer       \\
F6p  & Security-Measures               &  & F5d   & Agreement-not-specific        \\
F7p  & Brought-Tools                   &  & F10d  & Secrets-Disclosed-Outsiders   \\
F8p  & Competitive-Advantage           &  & F11d  & Vertical-Knowledge            \\
F12p & Outsider-Disclosures-Restricted &  & F16d  & Info-Reverse-Engineerable     \\
F13p & Noncompetition-Agreement        &  & F17d  & Info-Independently-Generated  \\
F14p & Restricted-Materials-Used       &  & F19d  & No-Security-Measures          \\
F15p & Unique-Product                  &  & F20d  & Info-Known-to-Competitors     \\
F18p & Identical-Products              &  & F23d  & Waiver-of-Confidentiality     \\
F21p & Knew-Info-Confidential          &  & F24d  & Info-Obtainable-Elsewhere     \\
F22p & Invasive-Techniques             &  & F25d  & Info-Reverse-Engineered       \\
F26p & Deception                       &  & F27d  & Disclosure-In-Public-Forum   
\end{tabular}
\end{table}

When presented with a new case a three ply argument is constructed. In the first ply a proponent cites the most-on-point precedent (i.e. the precedent with the greatest overlap of factors irrespective of which side they favour) decided for the side being argued for. Suppose this is the plaintiff. In the second ply the opponent either cites a counterexample (a case which favours the defendant and is at least as on point as the case cited by the plaintiff) or distinguishes the precedent by pointing to a factor favouring the plaintiff in the precedent but not the new case, or a factor favouring the defendant in the current case but not the precedent. In the third ply the plaintiff offers a rebuttal by distinguishing the counterexamples, or downplaying the distinguishing factor by pointing to a factor which can cancel the additional factor or a factor which can be substituted for the absent factor \cite{prakken2013formalization}.

CATO organised its factors into a \textit{factor hierarchy}. At the upper level are \textit{issues}, and below these are layers of abstract factors before the leaf nodes are reached. These leaf nodes are the base level factors shown in Table 1. The importance of this hierarchy is for determining whether distinctions can be downplayed: a factor can only be substituted for or cancel another factor if they have a common ancestor. The closer the ancestor the more persuasive the downplay. How persuasive the downplay is matters for prediction, where the success or otherwise of the rebuttal needs to be decided, but not for for explanation. The success or otherwise of the rebuttal is given by the outcome of the case, which shows whether or not the downplay was successful.

CATO's factor based model has inspired a number of formal accounts of precedential constraint \cite{horty2011reasons}, \cite{horty2012factor}, \cite{rigoni2015improved} and \cite{prakken2021}. These models are based on a way of representing precedents originating in \cite{prakken1998modelling}. Suppose we have a case with plaintiff factors $P$ and defendant factors $D$. Now the strongest reason to find for the plaintiff will be the conjunction of all the factors in $P$ and the strongest reason to find for the defendant the conjunction of all the factors in $D$. The outcome of the case will show which reason was preferred. A decision is taken to be constrained in these approaches if deciding for the other party would introduce an inconsistency into the set of preferences in the precedent base\footnote{In practice this formal notion of constraint may not actually be respected in a given judicial setting. For a jurisprudential discussion see \cite{schauer1987precedent}.}. Using all the factors available for the winning side is termed the \textit{results} model in \cite{horty2011reasons}. It may be, however, that a subset of the factors for the winning side is still sufficient to overcome the reason for the losing side. This would allow a subset of the winner's factors to be used in the preference. This is termed the \textit{reason} model in \cite{horty2011reasons}. A comparison of the two models is given in \cite{prakken2021}.

The reasoning in CATO: citation, distinguishing and counterexample, followed by rebuttal through downplaying distinctions and distinguishing counter examples was expressed as a set of argumentation schemes in \cite{wyner2007argument} and formalised in ASPIC+ in \cite{prakken2013formalization}. This formalisation uses the results model, and uses the full set of factors available to both sides. These schemes were proposed as a means of providing explanation for ML systems in \cite{prakken2021ac}. We will discuss the explanations from \cite{prakken2021ac} in the next section.

\section{Explanation with Argument Schemes}

\begin{figure} [t]
\center
\includegraphics[scale=0.575]{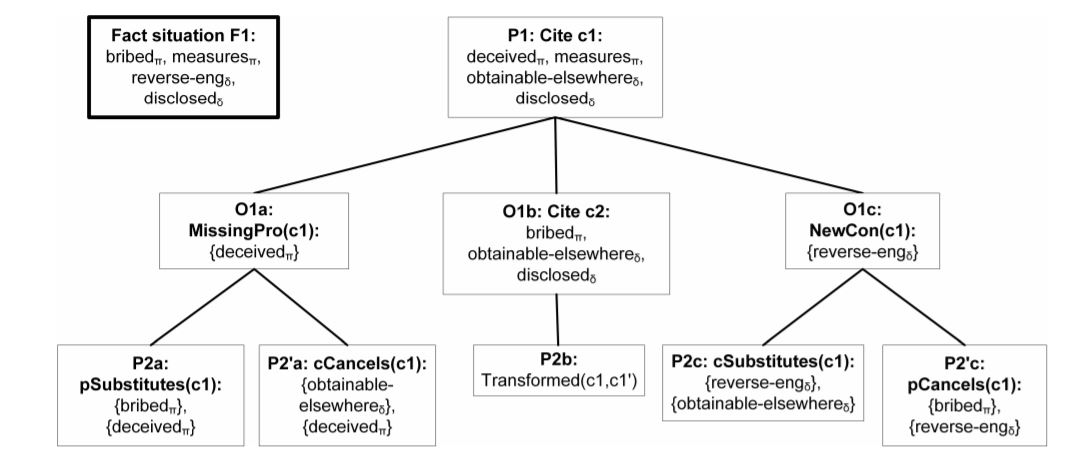} 
\caption{Example Dialogue Tree From \cite{prakken2021ac}} \label{PrakkenTree}
\end{figure}

Explanation in \cite{prakken2021ac} takes the form of a dialogue modelling the three ply argumentation structure of CATO. An example dialogue is shown in Figure~\ref{PrakkenTree}. Figure ~\ref{PrakkenTree} illustrates a particular example used in \cite{prakken2021ac}. The two precedent cases and the new case in that example are shown in Table~\ref{PrakkenCases}. The cases have been given mnemonic names. The top layer, put forward by the proponent, is an argument based on citing a precedent case. The second layer, objections by the opponent, comprises objections based on each of the two types of distinction ($O1a$ and $O1c$), and a counter example ($O1b$). The final layer shows the proponent's rebuttals: each objection is met by both a substitution and a cancellation ($P2a$ and $P2^\prime a$ for $O1a$ and $P2c$ and $P2^\prime c$ for $O1c$), and a rebuttal of the counterexample through a ``transformation'', which means that substitutions and cancellations can transform the case into a precedent for the proponent's side.

\begin{table}[h]
\center
\caption{Cases in the Example From \cite{prakken2021ac}} \label{PrakkenCases}
\begin{tabular}{llll}
Case        & Outcome & Plaintiff Factors & Defendant Factors \\
Deceived & P       & F6p F26p          & F10d F24d         \\
NoMeasures  & D       & F2p               & F10d F24d         \\
Bribed    & TBA       & F2p F6p           & F10d F16d        
\end{tabular}
\end{table}

The diagram in Figure~\ref{PrakkenTree} offers an explanation for a plaintiff win in \textit{Bribe}. \textit{Deceived} matches because the plaintiff took security measures (F6p) and disclosed information to outsiders (F10d). The defendant can now cite distinctions of both kinds: deception (F26p) was not used, and the information is re-engineerable (F16d) in the new case but not the precedent. Moreover \textit{NoMeasures} also matches on two factors and so is as on point as \textit{Deceived} and so can serve as a counter example. To counter $O1a$ the plaintiff now argues ($P2a$) that the lack of deception does not matter because bribery was used (F2p) and this can substitute for F26p. Alternatively it can be argued ($P2^\prime a$) that the additional defendant factor in the precedent, F24d, that the information was available elsewhere, cancels the additional plaintiff strength of the precedent coming from bribery. It is clear in this case that the substitution is more effective: bribery and deception play similar roles, both being different examples of the use of improper means. In the case of the additional defendant factor in the new case, F16d, used in $O1c$, it can be substituted by the additional factor in the precedent, F24d ($P2b$), or cancelled by the additional plaintiff factor, F2p ($P2^\prime b$). Again it seems that substitution is the better argument because of the similarity of the roles of F16d and F24d.

The fact that the substitutions are clearly better rebuttals than the cancellations (in this case: for other examples the reverse will be true) highlights a problem with the explanation. Although the arguments generated by the definitions are possible arguments in that they conform to the definitions of \cite{prakken2021ac}, they lack plausibility because they relate to entirely different concerns. That is, these objections fail to  make sense in domain terms. In CATO the strength of a downplay depended on how close the factors were in the factor hierarchy. In Figure~\ref{PrakkenTree}, however, it is not possible to tell which rebuttal succeeded, or whether a combination of the two was required. All the candidate arguments are presented, but it is the user that must supply the domain knowledge to assess how strong these arguments are, and which objections and rebuttals should succeed in the particular case. The reason for the decision is there, but the user must extract it. In order to guide the user, we turn to consider the knowledge of the structure of the domain represented by the use of issues.

\section{Issues}

\begin{figure}[t]
\center
\includegraphics[scale=0.495]{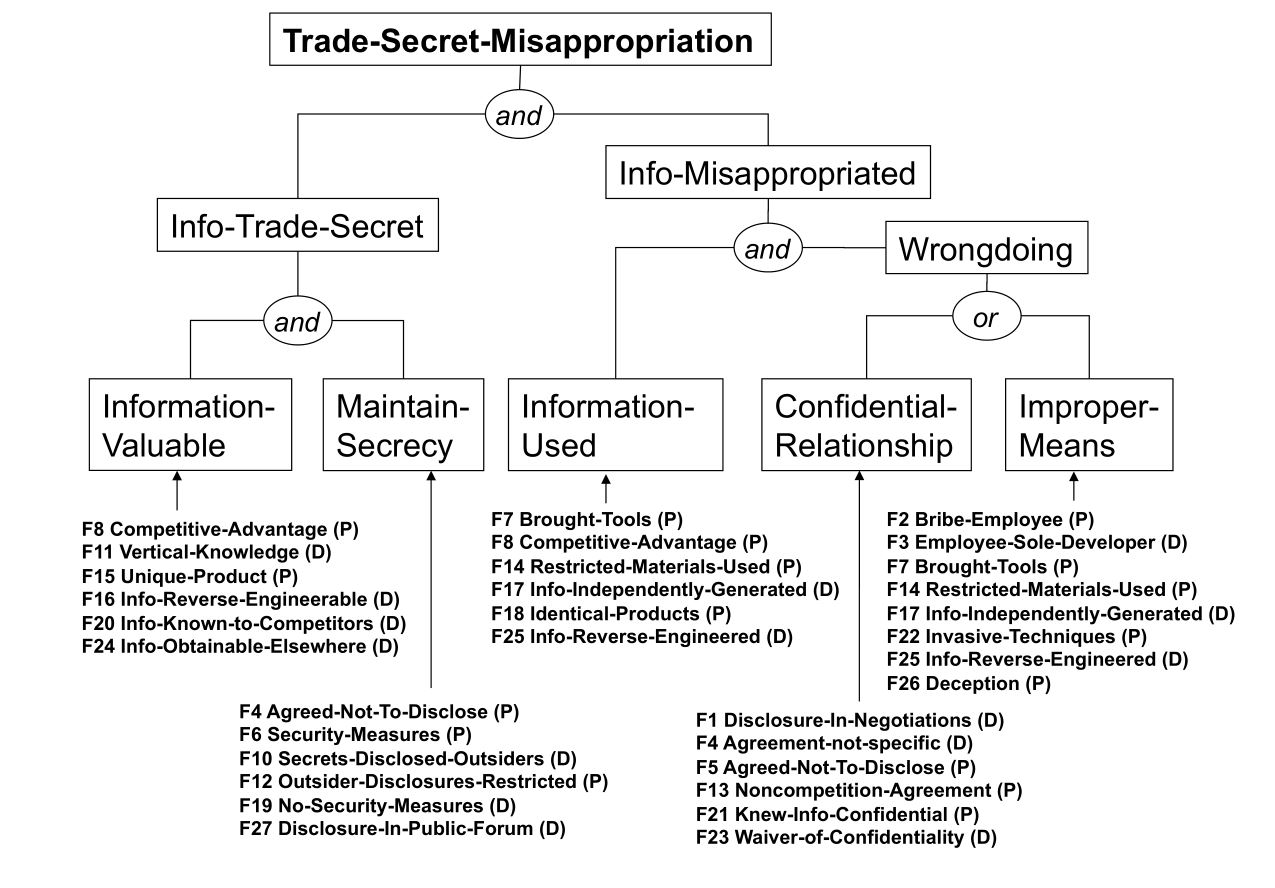} 
\caption{Issues and Factors From \cite{grabmair2016modeling}. Factors may relate to more than one issue.}\label{Grabmair} 
\end{figure}

In CATO the top level of the factor hierarchy was made up of \textit{issues}. The role of issues in a case is to identify the salient points that need to be shown in order to prove or defend a case, and, hence, the factors that are relevant to the different points. Issues in CATO served mainly to organise the explanation. CATO identified two main issues for Trade Secret misappropriation: the information had to be a trade secret (with the burden of proof on the defendant to show that it was not) and the information had to have been misappropriated (with the burden of proof on the plaintiff). Below these main issues were sub-issues. To be a trade secret the information had to be valuable and its secrecy adequately maintained. If the information was used, misappropriation could be shown either through a breach of confidence or through the use of improper means by the defendant. For a breach of confidence a confidential relation between plaintiff and defendant had to exist. Issues took on an additional significance when CATO was adapted to predict outcomes in the IBP system \cite{bruninghaus2003predicting} when the issues formed a top layer of necessary and sufficient conditions (termed the logical model in \cite{bruninghaus2003predicting}), with factor based reasoning used to determine the status of the leaf issues. This structure, strict logical rules at the top with case based reasoning to determine which rules applied, was earlier used in CABARET \cite{skalak1992arguments}, and was later adopted and adapted to accommodate his value judgement formalism by Grabmair \cite{grabmair2016modeling}. Figure~\ref{Grabmair} shows top level logical model and the allocation of factor to issues in \cite{grabmair2016modeling}\footnote{We use structure of the logical model in \cite{grabmair2016modeling} which differs slightly from that of \cite{bruninghaus2003predicting}.}.

We can now consider the example cases in Table~\ref{PrakkenCases} in terms of issues. We also include in Table~\ref{CasesByIssue} some other actual cases we will mention 
in this paper\footnote{Mason v. Jack Daniel Distillery, 518 So.2d 130 (Ala.Civ.App.1987), Leo Silfen, Inc. v. Cream, 29 N.Y.2d 387, 
Computer Print Systems v. Lewis, 422 A.2d 148 (1980), K \& G Oil Tool \& Service Co. v. G \& G Fishing Tool Serv., 314 S.W.2d 782, (1958),College Watercolor Group, Inc. v. William H. Newbauer, Inc., 468 Pa. 103, 360 A.2d 200 (1976), Arco Industries Corp. v. Chemcast Corp., 633 F.2d 435, 208 USPQ 190 (6th Cir.1980), E. V. Prentice Dryer Co. v. Northwest Dryer
\& Machinery Co., 246 Or. 78, 424 P.2d 227 (1967), Kinnear-Weed Corp. v. Humble Oil Refining Co. 150 F. Supp. 143 (E.D. Tex. 1956). Sheets v. Yamaha Motors Corp., USA, 657 F.Supp. 319 (1987), Commonwealth v. Robinson, 7 Mass.App.Ct. 470, 388 N.E.2d 705 (1979), MBL (USA) Corp. v. Diekman, 112 Ill.App.3d 229, 445 N.E.2d 418, 67 Ill.Dec. 938 (1983), A. H. Emery Co. v. Marcan Products Corporation, 380 F.2d 11 (1968), Ecologix, Inc. v. Fansteel, Inc., 676 F.Supp. 1374 (1988), Laser Industries, Ltd. v. Eder Instrument Co., 573 F.Supp. 987 (1983), Sandlin v. Johnson, 152 F.2d 8 (8th Cir.1945), Trandes Corp. v. Guy F. Atkinson Co., 996 F.2d 655 (4th Cir.1993), Ferranti Electric, Inc. v. Harwood, 43 Misc.2d 533, 251 N.Y.S.2d 612 (1964), The Boeing Company v. Sierracin Corporation, 108 Wash.2d 38, 738 P.2d 665 (1987). Note that the analysis into factors is mine, and for the purpose of illustration in this paper. It should not be relied on in a court of law.}. 

\begin{table}[t]
\caption{Cases with Factors Grouped by Issue.}\label{CasesByIssue}
\begin{tabular}{|l|l|l|l|l|l|l|}
\hline
\multirow{2}{*}{Case} & \multirow{2}{*}{P or D} & \multicolumn{2}{l|}{Trade Secret}                                                                                                & \multicolumn{3}{l|}{Misappropriation}                                                                                                                                                                  \\ \cline{3-7} 
                      &                          & \begin{tabular}[c]{@{}l@{}}Maintain\\    Secrecy\end{tabular} & \begin{tabular}[c]{@{}l@{}}Info\\    Valuable\end{tabular} & \begin{tabular}[c]{@{}l@{}}Improper\\    Means\end{tabular} & \begin{tabular}[c]{@{}l@{}}Confidential\\   Relationship\end{tabular} & \begin{tabular}[c]{@{}l@{}}Info\\   Used\end{tabular} \\ \hline
Deceived           & P                        & F6p F10d                                                         & F24d                                                          & F26p                                                           &                                                                           &                                                           \\ \hline
NoMeasures            & D                        & F10d                                                             & F24d                                                          & F2p                                                            &                                                                           &                                                           \\ \hline
Bribed     & ?     & F6p F10d     & F6p F16d         & F2p     &       &     \\ \hline
Mason     & P     &  F6p F15p    &  F6p  F16d        &        & F1d F21p      &    \\ \hline
Silfen    & D      &  F6p F11d        &    F6p               &        &        F21p     &      \\ \hline    
Lewis     & P      &F8p          &                      &        & F1d F21p  &         \\ \hline   
College   & P      &  F15        &                      & F26p    & F1d  &  \\ \hline    
Arco      & P      & F16d  F20d       &   F10d                   &          &      & \\ \hline
Sheets    & D &  &  F19d  F27d     &                     &                & F18p \\ \hline
Robinson    & D &  &  F10d F19d      &           F26p          &  F1d              & F18p \\ \hline
MBL    & D &  F6p F20d & F4p F5d F6p F10d     &                     &  F1d  F4p F5d F13p           &  \\ \hline
Prentice  &   D     &  F6p  F24d                &  F6p                        &  F3d        &      & \\ \hline
Kinnera-Weed & D   &  F6p  &   F6p                       &  F25d        & F21p     & F25d \\ \hline
Emery & P   &    &  F6p  F10d                       &          & F21p     & F18p \\ \hline
Laser & P   &  F6p  & F6p   F10d    F12p                   &          & F1d F21p     & F18p \\ \hline
Sandlin & D & F6p F16d   &  F6p F10d                       &          & F1d      &  \\ \hline
Ecologix & D &   &                        &          & F1d F21p  F23d   &  \\ \hline
Trandes & P & F6p  &  F4p F6p F10p F12p                      &  F1d F21p        &      &  \\ \hline
Ferranti & D & F20d  &                        &  F2p S17d       &      &  \\ \hline
Boeing & P & F15p  &  F10d  F12p                    &          &  F1d F21p    & F18d \\ \hline
\end{tabular}
\end{table}

\subsection{Explanation in CATO}

CATO organised its explanation in terms of issues. The explanation of the example in \cite{prakken2021ac} would be something like the following (adapted from the explanation of \textit{Mason} given in Figure 2.4 of \cite{aleven}). 

\begin{quote}
\textbf{Argument for Plaintiff in Bribed}.

Plaintiff should win a claim of trade secrets misappropriation.  Plaintiff’s information is a trade secret and defendant acquired plaintiff’s information through improper means.

\textbf{Plaintiff’s information is a trade secret.}

In \textit{Bribed,} plaintiff adopted security measures [F6p] 
This shows that plaintiff took efforts to maintain the secrecy of its information.

The fact that plaintiff disclosed its information to outsiders [F10d] does not preclude
a conclusion that plaintiff’s information is a trade secret. 
This is especially so where, as in \textit{Bribed}, plaintiff took security measures to protect
the information [F6p]. [\textit{Deceived}]

The fact that plaintiff’s information could be ascertained by examining or reverse engineering
plaintiff’s product [F16d] does not preclude a conclusion that plaintiff’s information is a trade secret [\textit{Mason}]. Moreover in \textit{Deceived}, the information was available elsewhere, which is not the case in \textit{Bribed}.

\textbf{Defendant acquired plaintiff’s information through improper means} by bribing an employee [F2p].

\textbf{\textit{NoMeasures} is not a counterexample} with respect to secrecy being maintained, since in that case, unlike \textit{Bribed}, the plaintiff did not take security measures to protect the secrecy of its information.
\end{quote} 

Organising by issues has several advantages. First it means that differences relating to uncontested issues are not considered: since the use of improper means is not contested, it does not matter which factor is used to establish it. This prunes the $O1a$ branch from the tree in Figure~\ref{PrakkenTree}. Second when downplaying the genuine distinction in $O1c$, it produces the correct rebuttal (P2c) because this is the factor related to the same issue, and ignores $P2^\prime c$. Also it explicitly cites the preference for F6p in past precedents as the reason for citing the precedent and rejecting the counterexample.

The preference for F6p over F16d is justified by reference to \textit{Mason}. Two points should be made here: the misappropriation in \textit{Mason} involved breach of confidence rather than improper means. This does not matter when using issues, but would provide a distinction for an approach without issues as used in \cite{horty2011reasons} and \cite{prakken2021}. This is true of the reason as well as the results model: F2p would have to be included in the reason for \textit{Mason}, since otherwise there would have been no breach of confidence. Using issues to organise our factors enables reasoning with portions of precedents \cite{branting1991reasoning}, which makes substantially more precedents available to make points. On the results model, even with issues, however, \textit{Mason} would be vulnerable to a distinction since it contains F15p whereas \textit{Bribed} does not, which provides an additional factor to outweigh F16d, suggesting that F6p might not be preferred to F16d on its own. If we consider the decision in Mason, however, we read:
\begin{quote}
We note that absolute secrecy is not required ...
for the recipe for Lynchburg Lemonade to constitute a trade secret — 
"a substantial element of secrecy is all that is necessary to provide trade secret protection." Drill Parts, 439 So.2d at 49.
The defendants also contend that Mason's recipe was not a trade secret because it could be easily duplicated by others. ... We do not think, however, that this evidence in and of itself could prevent such a conclusion. Rather, this evidence should be weighed and considered along with the evidence tending to show the existence of a trade secret.
In this regard, 
we note that courts have protected information as a trade secret despite evidence that such information could be easily duplicated by others competent in the given field. KFC Corp. v. Marion-Kay  Co., 620 F. Supp. 1160 (S.D.Ind. 1985); Sperry Rand Corp. v. Rothlein, 241 F. Supp. 549 (D.Conn. 1964).
\end{quote}

Here there is no reference to the uniqueness of the product (F15p), and so for the reason model we should take it that the decision in favour of Mason indicates that 
F6p is preferred to F16d on its own, without needing the support of F15p, and so \textit{Mason} is available for use in \textit{Bribed} when the reason model is applied at the issue level. Note, however, that if we are using the distribution of factors across issues in \cite{grabmair2016modeling} that was shown in Figure~\ref{Grabmair}, this would require us to consider the preference at the level of the \textit{TradeSecret} issue and  that InfoValuable is not required for the information to be a trade secret when F6p is present. We prefer instead to include F6p under InfoValuable as well as MaintainSecrecy: sufficient secrecy measures are considered enough to justify the information being deemed valuable. A factor can relate to more than one issue: for example F25d in Figure~\ref{Grabmair}. Putting F6p under InfoValuable to oppose F16d seems to accord with the decision in \textit{Mason} cited above. This enables us to continue to use  the logical model of \cite{bruninghaus2003predicting}  and \cite{grabmair2016modeling} which requires \textit{Bribed} to show both that the information was valuable and that efforts to maintain secrecy had been taken. Without F6p being included under InfoValuable this model would fail since that there is no factor to contest the claim that
F16d would mean that the information lacked value.

\subsection{The Centrality of Issues}

From the above discussion we see that organising explanations around issues provides focus and enables irrelevant factors and uncontested issues to be ignored, so avoiding swamping the recipient of the explanation with an excess of information. This accords with the widespread popularity of the Issue-Rule-Application-Conclusion (IRAC) methodology of legal analysis. IRAC is widely taught in law schools\footnote{For example, City University of New York (https://www.law.cuny.edu/legal-writing/students/irac-crracc/irac-crracc-1/) and  Elizabeth Haub School of Law at Pace University (https://academicsupport.blogs.pace.edu/2012/10/26/the-case-of-the-missing-a-in-law-school-you-cant-get-an-a-without-an-a/). Use of IRAC is advocated by the LexisNexis survival guide for law students available at https://www.lexisnexis.co.uk/students/law/.}, although often there are variants which include an additional item or reorder the items, perhaps beginning with the conclusion. IRAC was advocated for the explanation of outcomes from factor based reasoning in \cite{bench2020explaining}.  
"Issue spotting" has a long history in AI and Law, dating back to the work of Gardner \cite{gardner} and puesued by Gordon \cite{gordon1989issue} and \cite{gordon1993pleadings}. 

The question arises, however, as to what should be counted as an issue. Issues could be very coarse grained such as \textit{was the information a trade secret?}, or relate to the fine grained abstract factors of the CATO factor hierarchy, such as \textsl{did the plaintiff take adequate security measures with respect to the defendant?}. In the next section we will offer a tree of issues in the form of an Abstract Dialectical Framework (ADF) \cite{brewka2010abstract} as used for the representation of legal knowledge in \cite{al2016methodology}.

\begin{table}[t]
\center
\caption{ADF for Trade Secret Misapproriation}\label{ADF}
\tiny
\begin{tabular}{|l|l|l|l|}
\hline
Node                        & Children                                                                                                              & Acceptance   Conditions    & Justification                                                                                                                                                   \\ \hline
\begin{tabular}[c]{@{}l@{}}Trade\\     Secret\\     Misappropriation\end{tabular}  & \begin{tabular}[c]{@{}l@{}}InfoTradeSecret  \\ InfoMisappropriated\end{tabular}                                   & \begin{tabular}[c]{@{}l@{}}ACCEPT   IF  InfoTradeSecret    \\  \hspace{3em}  AND     InfoMisappropriated  \\ REJECT\end{tabular}                                                            &   Restatement   \\ \hline
InfoTradeSecret             & \begin{tabular}[c]{@{}l@{}}InfoValuable   \\ MaintainSecrecy\end{tabular}                                          & \begin{tabular}[c]{@{}l@{}}ACCEPT  IF InfoValuable \\  \hspace{3em} AND  MaintainSecrecy\\     REJECT\end{tabular}                                                              &   Restatement                  \\ \hline
\begin{tabular}[c]{@{}l@{}}Info \\    Misappropriated\end{tabular}         & \begin{tabular}[c]{@{}l@{}}WrongDoing\\    InfoUsed\end{tabular}                                                   & \begin{tabular}[c]{@{}l@{}}ACCEPT IF WrongDoing   \\ \hspace{3em} AND InfoUsed\\     REJECT\end{tabular}                  &   Restatement                                           \\ \hline
InfoValuable                & \begin{tabular}[c]{@{}l@{}}F6p\\     F8p\\     F11d\\ F15p \\   InfoObtainable\end{tabular}                           & \begin{tabular}[c]{@{}l@{}}  REJECT IF F11d\\   ACCEPT IF F8p\\       ACCEPT IF F6p\\   ACCEPT IF F15p \\REJECT IF   InfoObtainable\\     ACCEPT\end{tabular} & \begin{tabular}[c]{@{}l@{}}Silfen \\     Lewis \\     Mason \\ College  \\ Restatement \\ \\ \end{tabular}                               \\ \hline
MaintainSecrecy             & \begin{tabular}[c]{@{}l@{}}F27d\\    F6p \\ F19d\\     MeasuresOutsiders\end{tabular}   & \begin{tabular}[c]{@{}l@{}}REJECT IF F27d\\    REJECT IF F19d \\ ACCEPT IF F6p\\     REJECT IF NOT\\ \hspace{3em}  MeasuresOutsiders\\     ACCEPT\end{tabular}     &    \begin{tabular}[c]{@{}l@{}}Sheets \\     Robinson \\     Emery \\ \\ Restatement  \\ \\ \end{tabular}                               \\ \hline
WrongDoing                  & \begin{tabular}[c]{@{}l@{}}F3d\\     OwnEfforts\\    ImproperMeans\\     ConfidRelation\end{tabular} & \begin{tabular}[c]{@{}l@{}}REJECT IF F3d\\     REJECT IF   OwnEfforts\\     ACCEPT IF   ImproperMeans\\     ACCEPT IF   ConfidRelation\\     REJECT\end{tabular}  &    \begin{tabular}[c]{@{}l@{}}Prentice \\     Restatement \\     Restatement \\ Restatement  \\ \\ \end{tabular} \\ \hline
ImproperMeans               & \begin{tabular}[c]{@{}l@{}}InfoMisuse\\     IllegalAct\end{tabular}                                                 & \begin{tabular}[c]{@{}l@{}}ACCEPT IF   InfoMisuse\\     ACCEPT IF   IllegalAct\\     REJECT\end{tabular}  &    \begin{tabular}[c]{@{}l@{}}Restatement \\ Restatement  \\ \\ \end{tabular}                                                                     \\ \hline
ConfidRelation    & \begin{tabular}[c]{@{}l@{}}NoticeConfid   \\     ExplicitAgreement\end{tabular}                          & \begin{tabular}[c]{@{}l@{}}ACCEPT IF   NoticeConfid    \\     ACCEPT IF   ExplicitAgreement\\     REJECT\end{tabular}  &    \begin{tabular}[c]{@{}l@{}}Restatement \\ Restatement  \\ \\ \end{tabular}                                              \\ \hline
InfoUsed                    & \begin{tabular}[c]{@{}l@{}}InfoMisue\\     OwnEfforts\\    F8p\\     F18p\end{tabular}                         & \begin{tabular}[c]{@{}l@{}}ACCEPT IF   InfoMisue\\  ACCEPT IF F8p\\     ACCEPT IF   F18p\\     REJECT  IF OwnEfforts\\     ACCEPT\end{tabular}   & \begin{tabular}[c]{@{}l@{}}Restatement \\     Lewis \\     Emery \\ Restatement  \\ \\ \end{tabular}                       \\ \hline
InfoMisuse                  & F7p F14p                                                                                                              & ACCEPT IF F7p   OR F14p            & Restatement                                                                               \\ \hline
IllegalAct                  & F2p F22p F26p                                                                                                            & \begin{tabular}[c]{@{}l@{}}ACCEPT if   F2p\\  \hspace{3em}   OR F22p  OR F26p \\     ACCEPT\end{tabular}    & Restatement                                                                                                                                                   \\ \hline
NoticeConfid     & F1d F21p F23d                                                                                                         & \begin{tabular}[c]{@{}l@{}}REJECT if   F23d\\     ACCEPT IF   F21p\\     REJECT IF   F1d\\     ACCEPT\end{tabular}   &    \begin{tabular}[c]{@{}l@{}}Ecologix \\    Laser \\     Sandlin   \\  \\ \end{tabular}                                                       \\ \hline
ExplicitAgreement           & F4p F5d F13p                                                                                                          & \begin{tabular}[c]{@{}l@{}}REJECT IF F5d\\     ACCEPT IF F4p   OR F13p\\     REJECT\end{tabular}  & \begin{tabular}[c]{@{}l@{}}MBL \\ Trandes  \\ \\ \end{tabular}                                                                             \\ \hline
InfoObtainable              & \begin{tabular}[c]{@{}l@{}}F15p F16d \\    F20d F24d\end{tabular}                                                                                                    & \begin{tabular}[c]{@{}l@{}}REJECT IF   F15p\\     ACCEPT IF   F24d OR F20d\\     REJECT\end{tabular}    & \begin{tabular}[c]{@{}l@{}}College \\ Ferranti  \\ \\ \end{tabular}                                                                      
 \\ \hline                                                                                      
MeasuresOutsiders        & F10d F12p                                                                                                             & \begin{tabular}[c]{@{}l@{}}ACCEPT IF   F12p\\     REJECT IF   F10d\\     ACCEPT\end{tabular}     & \begin{tabular}[c]{@{}l@{}}Trandes \\ Arco  \\ \\ \end{tabular}                                                                               \\ \hline
OwnEfforts                  & F17d F25d                                                                                                             & \begin{tabular}[c]{@{}l@{}}ACCEPT IF   F17d OR F25d\\ REJECT\end{tabular}       & \begin{tabular}[c]{@{}l@{}}Kinnear-Weed \\ \\ \end{tabular}                                                                                                    \\ \hline
\end{tabular}
\end{table}
\section{An ADF of Issues}

Table~\ref{ADF} presents the issues used to decide questions of Trade Secret Misappropriation in the form of an ADF. The decomposition of InfoMisappropriated follows \cite{grabmair2016modeling} as shown in Figure~\ref{Grabmair} rather than \cite{bruninghaus2003predicting}. Traditionally models of reasoning with legal precedents include the \textit{rule} model and the \textit{balance of factors} model \cite{varsava2018realize}. Although CATO style reasoning relflects the balance of factors model, the ADF in Table~\ref{ADF} uses the rule model in all the acceptance conditions, assuming that there are sufficient precedents to justify the priorities in every node. In Table~\ref{ADF} the fourth column shows the justification for the acceptance condition: either the logical model implied by the \textit{Restatement} or a precedent which expressed a particular preference\footnote{Because the use of issues mean that at most five factors need to be considered for any given node, it is feasible to envisage enough precedents to resolve the node. This would not be so as the whole case level, where there are $2^{26}$ possible models.}. 

However, following the examples of CABARET \cite{skalak1992arguments} and Issue Based Prediction (IBP) \cite{bruninghaus2003predicting}, at some point in each branch the rule model might not be appropriate and so the balance of factors model would be required. This will be particularly so in the early stages of the development of a body of case law, when there will not yet be sufficient precedents to determine all the required preferences. For example, F18p (Unique Product) may be a strong indication that the information was used, but might not be as decisive as suggested in the ADF in Table~\ref{ADF}. If we use a balance of factors model we can choose between balancing the factors representing the children of the node in question or the factors which result from unfolding the nodes below the node concerned. Thus InfoTradeSecret could be resolved by regarding InfoValuable and InfoTradeSecret as a conjunction so that both must be true (rule model); or allowing some trade off between them (coarse grained balance of factors), or balancing the relevant base level factors \{F6p, F8p, F10d, F12p, F11d, F15p, F16d, F19d, F20d, F24d, F27d\} (fine grained balance of factors). Also perhaps a mixed granularity could be used, grouping some of the base level factors into abstract factors, e.g. F10d and F12p could be considered together as MeasuresOutsiders. It would be an interesting exercise to see what effect these different granularities might have, but for this paper we will assume a mature domain allowing us to use a pure rule model.

\subsection{Using Issues for Explanation}

Now we have the ADF, we can identify the issues in particular cases. The issues in a case will be the lowest nodes spanning both a pro-plaintiff factor and a pro-defendant factor. We illustrate this principle by applying it to the cases in Table~\ref{PrakkenCases}:

In \textit{Deceived} MaintainSecrecy is an issue because Security Measures (F6p) favours the plaintiff whereas MeasuresOutsiders favours the defendant (F10d).
\textit{Deceived} was found for the plaintiff, and so  MeasuresOutsiders $\prec$ F6p. We could use \textit{Emery} as a precedent to justify this preference. InfoValuable is also an issue because it contains both 
F6p and F24d. Again since the plaintiff won we can infer that InfoAvailableElsewhere $\prec$ F6p, a preference justified by \textit{Mason}. 

In IRAC terms: 
\begin{quote}
An issue is whether that there were no efforts to maintain secrecy with respect to outsiders means that secrecy was not maintained. The rule is that Not MeasuresOutsiders $\prec$ F6p (\textit{Emery}). Application is that  Not MeasuresOutsiders applies because F10d is present, but F6p is also present. Therefore secrecy was maintained. A second issue is whether that the information was obtainable elsewhere means that the information was not valuable. The rule is InfoAvailableElsewhere $\prec$ F6p (\textit{Mason}). The rule applies because F24d establishes InfoAvailableElsewhere and F6 is present. Therefore, the information was valuable.
\end{quote}

In \textit{NoMeasures} only the root node, TradeSecretMisappropriation, spans contested factors. Here the defendant can establish that the information is not a trade secret because it wins both branches for that issue and so the defendant is found for.

In IRAC terms:
\begin{quote}
The issue is whether a Trade Secret was misappropriated, when the information was misappropriated but not a trade secret. The rule is if not InfoTradeSecret then not TradeSecretMisappropriation (\textit{Restatement of Torts}). The rule applies because the information was obtainable elsewhere because F24d was present (\textit{Ferranti}) and the efforts to maintain secrecy with respect to outsiders were inadequate  because F10d is present (\textit{Arco}). Therefore there was no TradeSecretMisappropriation.
\end{quote}
In \textit{Bribed} we have a situation similar to \textit{Deceived} except that InfoAvailableElsewhere is established by F16d rather than F24d. The IRAC explanation is thus similar.
\begin{quote}
An issue is whether in there were inadequate efforts to maintain secrecy with respect to outsiders means that secrecy was not maintained. The rule is that MeasuresOutsiders $\prec$ F6p (cf \textit{Deceived}). Application is that  MeasuresOutsiders applies because F10d is present, but F6p is also present. Therefore secrecy was maintained. A second issue is whether that the information was obtainable elsewhere means that the information was not valuable. The rules is InfoAvailableElsewhere $\prec$ F6p (\textit{Mason}). The rule applies because F16d establishes InfoAvailableElsewhere and F6 is present. Therefore, the information was valuable.
\end{quote}

In \textit{Mason} there are two issues. The first, InfoValuable, is the what sets our precedent for InfoAvailableElsewhere $\prec$ F6p.  The second relates to whether there was notice of confidentiality with both F1d and F21p present. The answer is that there was because F1d $\prec$ F21p, established in \textit{Laser}.

In IRAC terms:
\begin{quote}
An Issue is whether that the information was obtainable elsewhere means that the information was not valuable. The rule used by this court is InfoAvailableElsewhere $\prec$ F6p. The rule applies because F26d establishes InfoAvailableElsewhere and F6 is present. Therefore, the information was valuable.
A second issue is whether there was notice of confidentiality where information was disclosed in negotiations and the defendant knew the information to be confidential. The rule is F1d $\prec$ F21p (\textit{Laser}). The rule applies because F1d and F21p are present. Therefore there was notice of confidentiality.
\end{quote}

Whereas the explanations from \cite{aleven} and \cite{prakken2021ac} begin at the top level and work down to the decisive facts, the IRAC explanations begin with the decisive facts. The IRAC explanations are thus very focused and do not aspire to give an exhaustive account, dotting every `i' and crossing every 't', but instead home in on what mattered in the particular case. As such they assume that the person to whom the explanation will have some knowledge of the domain. A person familiar withe logical model in Figure~\ref{Grabmair} or, better yet, the nodes of the ADF in Table~\ref{ADF} will have no difficulty in seeing why these points matter and how they decide the case. This suppression of shared background to highlight the decisive considerations was one of the original motivations for argument based explanation 
\cite{bench1991argument}. For those who need or want a fuller explanation, a dialogue seeking summary information in the manner of \cite{bench1993argument} can be initiated. Thus asking \textit{SO?} will display the parent node of the issue and asking \textit{WHY}? will display a child node. 

Thus for \textit{Bribed} asking \textit{SO?} for the first issue will produce \textit{Secrecy was Maintained}. Asking \textit{SO?} again will produce T\textit{he Information was a Trade Secret}. A third \textit{SO?} will produce \textit{The Trade Secret was Misppropriated}. Now a series of \textit{WHY?}s will produce \textit{The Information Was Misapproriated}, \textit{There Was Wrongoing}, \textit{There was an Illegal Act} and finally \textit{The Information Was Obtained by Deception}. Of course the users may stop the flow of information when they have enough to see the correctness of the solution.

\subsection{The Boeing Company v. Sierracin Corporation, 108 Wash.2d 38, 738 P.2d 665 (1987)}

We will now give a full example, using \textit{Boeing}\footnote{The description of the case is based on the opinion of Justice Dore, available at https://casetext.com/case/boeing-company-v-sierracin-corporation.}. Boeing is a very well known aircraft manufacturer. They do not, however, make all the parts themselves, but sub-contract certain components. One such component was the 7/7/7 cockpit windows. Five of these windows lie on each side of the aircraft's nose, each performing multiple critical functions such as defogging, withstanding cabin pressurisation, and providing a clear range of vision. The three major suppliers of aircraft windows in the United States are PPG Industries, Inc.; Swedlow, Inc.; and Sierracin. Sierracin had supplied Boeing with other products for many years. Boeing's drawings for the 7/7/7 cockpit windows are unique, detailed blueprints containing approximately 500 critical tolerances, dimensions, specifications and material requirements.  Boeing helped Sierracin enter the 7/7/7 window market by providing Sierracin with FAA authorized drawings, technical assistance and tooling, and by awarding it contracts for some of Boeing's 7/7/7 window needs in 1982 and 1983.

In 1984 after alleged breaches of contract because of late deliveries of windows, Boeing chose not to renew contracts with Sierracin, and instead signed a 5-year 100 percent requirements contract with PPG Industries, Inc. Sierracin decided, however, to continue manufacturing windows for  sale on its own in the 7/7/7 ``after market'' 
(i.e., spare parts market). As a supplier, Sierracin received Boeing's requests for quotations, which provided that all orders were subject to its confidential terms and conditions. Boeing alleged that Sierracin misappropriated its trade secrets concerning the design of aeroplane windows.

Sierracin signed over 270 contracts with Boeing, each containing the following language:
``Confidential Disclosure. Seller shall keep confidential . . . all proprietary information''.

The defence was that the information was not a trade secret because it had been disclosed to outsiders (F10d) and had not been misappropriated because it had been disclosed to Sierracin in negotiations (F1d). This was countered by the explicit confidentiality agreement that Boeing required suppliers to sign (F12p), and that Sierracin were well aware of the confidential nature of the information (F21p). The court found for Boeing.

The explanation based on the above proposal would look like this:

\begin{quote}
\textbf{The Boeing Company v. Sierracin Corporation, 108 Wash.2d 38, 738 P.2d 665 (1987)}.

The decision is for the plaintiff. There are two issues:
\begin{enumerate}
    \item Whether adequate measures with respect to outsiders were taken when the information was disclosed to outsiders, but these disclosures were restricted. The rule is Secrets-Disclosed-Outsiders $\prec$ Outsider-Disclosures-Restricted (Trandes Corp. v. Guy F. Atkinson Co., 996 F.2d 655 (4th Cir.1993)). The rule applies because F10d and F12p are present. Therefore, adequate measures with respect to outsiders were taken.
    \item Whether there was notice of confidentiality when the information was disclosed in negotiations, but the defendant knew that the information was confidential. The rule is Disclosure-In-Negotiations $\prec$ Knew-Info-Confidential (Laser Industries,Ltd. v. Eder Instrument Co., 573 F.Supp. 987 (1983)). The rule applies because F1d and F21p are present. Therefore, there was notice of confidentiality,
\end{enumerate}
\end{quote}

The user may now interrogate the system further to help in understanding why these issues matter.
\begin{quote}
   \textit{Issue 1: So?}
   
   Reply 1: Secrecy was Maintained (Restatement of Torts section 757, comment(b), bullet 3).
   
   \textit{Reply 1: So?}
   
   Reply 2: The information was a Trade Secret (Restatement of Torts section 757, comment(b).
   
   \textit{Reply 2: Why?}
   
   Reply 3: The information was valuable. (Restatement of Torts section 757, comment(b), bullet 3).
   
   \textit{Reply 3: Why?}
   
   The issue was unopposed. Further the product was unique (F15p).
   \end{quote}
   
   Satisfied as to Issue 1, the user now turns to issue 2.
   \begin{quote}
       
   \textit{Issue 2: So?}
   
   Reply 4: There was a Confidential Relationship (Restatement of Torts, section 757(b).
   
   \textit{Reply 4: So?}
   
   Reply 5: The Information Was Misappropriated (Restatement of Torts, Section 757, General Principle).
   
   \textit{OK}
   \end{quote}
   
   Since the user knows that the plaintiff should win if the information was a trade secret and misappropriated, the dialogue is terminated here.

    \section{Discussion}
    
    In this section we will consider three points: the quality of the explanations; the implications for accounts of precedential constraint; and how to accommodate dimensional facts.
    
    \subsection{Quality of Explanation}
    
    In his illuminating survey on explanation \cite{miller2018explanation}, Miller identifies four features of good explanations. These are:
\begin{itemize} 
\item Explanations are \textit{contrastive}. As well as explaining why a particular classification is appropriate, a good explanation will also say why other classifications are not, often using counterfactuals and hypotheticals. 
\item Explanations are \textit{selective}. Rarely is a logically complete explanation provided, but rather only the most salient points are presented unless more detail is required by the recipient of the explanation. The assumption is that there will be a considerable degree of shared background knowledge, and so the explanation need only point to some fact or rule as yet unknown to the recipient.
\item Explanations are rarely in terms of \textit{probabilities}.  Using statistical generalisations to explain why events occur is unsatisfying since they do not explain the generalisation itself. Moreover, the explanation typically applies to a single case, and so would require some explanation of why that particular case is typical.
\item Explanations are \textit{social}. Explanations involve a transfer of knowledge, between particular people in a particular situation and so are relative to the explainer’s  beliefs about the explainee’s beliefs.
\end{itemize}
    
  The explanations produced by using issues as described above have these features. The explanation is contrastive because it begins with the issue which will include a factor for the other side, and so suggests why the decision would have gone otherwise had the preferred factor for the winning side not been present. They are selective because they begin by stating the particular reason for the decision, and offers any further explication of the background knowledge explaining exactly why this matters only on request. Like most explanations in law, no probabilities are used: specific precedents are used to justify the rules rather than some degree of support as in, for example, association rule mining \cite{wardeh2009padua}. Finally the explanation is social in that it adopts a form of explanation (IRAC) that is widely used in the legal community which transfers  knowledge in a way tailored to the situation and the user. The particular strength of the proposed method is perhaps its \textit{selectivity}, which contrasts with the dialogue proposed in \cite{prakken2021ac} which included all available arguments and objections, and the explanations in \cite{aleven} and \cite{al2016methodology} which again covered every aspect without regard as to what the user might already know. 

\subsection{Implications for Precedential Constraint}

The increased effectiveness of formal characterisations of precedential constraint when applied at the issue level rather than at the whole case level was discussed in \cite{bench2021}. A problem with the results model was that, given, the large, number of factors, there are so many possible case descriptions ($2^{26}$) that it is all too easy to avoid the constraint of a precedent by pointing to a distinguishing factor. The use of the reason model alleviates this problem to some extent, but far from completely, as illustrated in \cite{bench2021}. However, the number of possible case descriptions at the \textit{issue} level is greatly reduced. Examination of Table~\ref{ADF} shows that no node has more than five children and so no node would require more than 32 cases to be fully resolved. Many of the nodes would require fewer: MeasuresOutsiders, measures for example has only two children, so that there are only four distinct case descriptions. Actually only three of these are possible, since F12p (restrictions placed on outsider disclosures) cannot be present without F10d (disclosures to outsiders). Also some precedents can be used for several nodes: the cases of MeasuresOutsiders required in MaintainSecrecy can also be used to resolve MeasuresOutsiders itself. This being so, fewer than 150 precedents would be required to resolve the tree completely, even on the results model. This number could be substantially reduced by applying the reason model to the issues. Since IBP \cite{bruninghaus2003predicting} used over 180 cases, it would seem that this might be a feasible way of addressing the problem. Moreover, various Machine Learning approaches use even bigger datasets: over 15,000 were available to \cite{branting}. If we have a dataset already available in a form in which factors can be straightforwardly ascribed - as might be expected when decisions are made on the basis of an application form, which is common in fields such as welfare benefits - or we have a machine learning program which ascribes factors as in \cite{branting}, we can apply this approach if we can associate these factors with issues. This should enable us to learn the acceptance conditions for the nodes using a variety of ML techniques, including such traditional techniques as rule induction \cite{movzina2005argument}.

The above suggests that whether we are using precedents for explanation or for learning, they are best considered in terms of issues rather than as whole cases.

\subsection{Factors With Magnitude}

The above presents a rather sanguine view of the possibility of predicting legal decisions, suggesting that a couple of hundred cases should enable the identification of a set of rules that would give complete accuracy. Leaving aside the many problems with any approach predicting legal decisions on the basis of a set of past cases, including that the law is constantly evolving so that predictions based on old data become unreliable \cite{medvedeva2019using} and that any collection of precedents is likely to contain decisions that were biased or incorrect, there is another serious problem. The above has assumed that, as in \cite{aleven}, factors are either present or absent, and can be ascribed to cases unequivocally. In practice, however, this is not so. Consider the Restatement of Torts:

\begin{quote}
Some factors to be considered in determining whether given information is one's trade secret are: (1) the \textbf{extent} to which the information is known outside of his business; (2) the \textbf{extent} to which it is known by employees and others involved in his business; (3) the \textbf{extent} of measures taken by him to guard the secrecy of the information; (4) the \textbf{value} of the information to him and to his competitors; (5) the \textbf{amount} of effort or money expended by him in developing the information; (6) the \textbf{ease or difficulty} with which the information could be properly acquired or duplicated by others.  \textbf{Emphasis mine}.
\end{quote}

From this it is clear that many of the aspects are not simply present or absent, but are present or absent to some degree and so require some judgement as to whether they were present to a degree \textit{sufficient} to permit the ascription of the factor. It should be remembered that the factors used in CATO derive from the dimensions proposed in HYPO (\cite{rissland1987case} and \cite{ashley1991modeling}). Dimensions were aspects of a case which could, if applicable, take a range of values which would increasingly favour a particular party.  The relationship between dimensions and factors is discussed in \cite{rissland2002note} and \cite{bench2021}. In fact, in HYPO, ten of the thirteen dimensions could take only two values and so either were inapplicable or favoured one of the parties. For example, the bribery dimension either favoured the plaintiff if bribery had taken place, or was inapplicable. It therefore maps straightforwardly to a single factor, F2p. Three of the dimensions did, however, span a rage of values. Competitive Advantage was either inapplicable, neutral, or favoured the plaintiff, mapping to  F8p. Disclosures to Outsiders was inapplicable if there had been no disclosures, or favoured the defendant if there had been sufficient disclosures (F10d) and neutral otherwise. Note, however, that extreme pro-defendant values on this dimension gave rise to the the more powerful factor F27d when the information was considered in the public domain. This is important because a plaintiff factor might be preferred to F10d, but not F27d. This is true of F6p in Table~\ref{ADF}. The most interesting dimension is security measures. At one end this favours the defendant and so maps into F19d, whereas at the other it maps into F6p and favours the plaintiff. It is thus always applicable. Many cases, however, contain neither F19d nor F6p, suggesting that the middle of the range is neutral so that no factor is applicable. Note that where a dimension can favour both sides, two distinct factors are used: this is because in general the absence of factor is  not a reason to decide for the other side. The was explained by Rissland and Ashley in \cite{rissland2002note}:
\begin{quote}
[...] the Security-Measures
dimension was broken into two factors: Security-Measures [F6p], favoring the plaintiff,
and No-Security-Measures [F19d], favoring the defendant. This was done because judges
explicitly said that the fact that plaintiff had taken no security measures was a
positive strength for the opponent. By contrast, Ashley and Aleven did not create a “No-Secrets-Disclosed-Outsiders” factor because they found no cases where
judges had said that the absence of any disclosures to outsiders was a positive
strength for the plaintiff. (\cite{rissland2002note}, p 69).
\end{quote}

Most of the CATO factors derive from two valued dimensions and so can be considered either present or absent, and so do not require special consideration. The three factors deriving from dimensions with ranges of value, however, do need something more. Horty in \cite{horty2017reasoning} and \cite{horty2019reasoning} introduced the notion of \textit{factors with magnitude}, and discussed how these could be accommodated in a theory of precedential constraint. Rigoni addressed this problem in \cite{rigoni2018representing}, and Horty modified his approach in \cite{horty2021}. A comparison of their approaches is given in \cite{prakken2021ac}. Although this was taken in \cite{prakken2021ac} to imply that precedential constraint should be expressed in terms of dimensions rather than factors, it was argued in \cite{bench2021} that this is perhaps not the best approach. We will explain how we can accommodate dimensions in an account of precedential constraint based on factors.

Many have seen reasoning with legal cases as a two stage process: first factors are assigned on the basis of the facts in a cases, and then these factors are considered in the the light of precedent cases to see whether the decision is constrained. This two stage approach is described in \cite{prakken2013formalization}

\begin{quote}
Once the facts of a case have been established - and this is rarely straightforward
since the move from evidence to facts is often itself the subject of debate - legal reasoning can be seen, following Ross \cite{ross1957tu} and Lindhal and Odelstad \cite{cdsv2006open}, as a two stage process,
first from the established facts to intermediate predicates, and then from these intermediate predicates to legal consequences. CATO has been explicitly identified with
the second of these steps (e.g. \cite{bruninghaus2003predicting}). (\cite{prakken2013formalization}, p 22). 
\end{quote}

This approach has been used not only in \cite{prakken2013formalization}, but also in \cite{ashley2009automatically}
and further advocated in \cite{branting} and \cite{branting2020}.

If we adopt this two stage model, we can see factors with magnitude as factors deriving from a dimension with more than two values. It may, like Security Measures, favour either side, with a neutral area in which no factor is applicable, or, like disclosures, give rise to two factors favouring the same side with different strengths, as with F10p and F27p.
Now we must determine which factor, if any, should be ascribed in a particular case given that it lies at a certain point on the range (termed by Horty a \textit{dimensional fact}). This can be determined by precedents using either the results model or the reason model \cite{horty2019reasoning}. Rigoni proposed that precedents should be regarded as identifying \textit{switching points} on the dimensions, the points at which factors come to be, and cease to be, applicable. The fact that whether a factor is applicable or not is itself be debatable,  complicates the first stage of the process. However, once the set of applicable factors has been identified, the second stage can proceed as described above.

Another, perhaps more serious, problem is that these factors may not be independent. For example, cases arise which require balancing the interests of the state in enabling the enforcement of laws with the privacy interests of its citizens \cite{bench2011argument}. In such cases it may seem necessary to trade off one factor against another, so that the more serious the suspected crime the greater the intrusion on privacy that is justified. Such balancing of interests has been discussed in \cite{lauritsen2015balance} and \cite{gordon2016formalizing}. We would, however, in line with the two stage approach, follow \cite{Bench-CaponA17} and see the question not as a balance between factors, but as a question of the ascription of factors on the basis of dimensional facts, so that factors themselves can continue to be seen as independent.


The example in \cite{horty2019reasoning} concerns change of fiscal domicile. Among other things to be considered are the length of stay abroad and the percentage of income earned abroad. The longer the absence and the greater the amount, the more change is favoured. We may now have a decision where an absence of 36 months and earnings of 60\% favoured  change, while an absence of 48 months but only 20\% earnings favoured no change. This suggests that absence and income are not independent, but trade off against each other. Suppose we have a third case also with 20\% earnings, but an absence of 60 months, further indicating the existence of a trade off. 

\begin{figure} [t]
\centering
\includegraphics[scale=0.85]{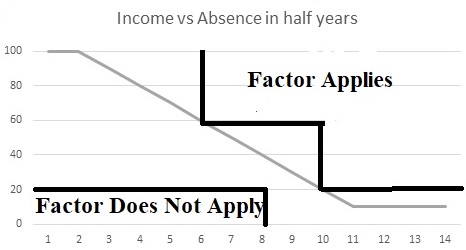} 
\caption{Trade off between absence and income} \label{Trade}
\end{figure}

The suggestion in \cite{Bench-CaponA17} is to introduce a factor ascribed to the case on the basis of the dimensional \textit{facts} recording absence and income. In this case a suitable factor would  be IncomeSufficientGivenAbsence and would favour change.
Each precedent for change will block off an area where the factor definitely applies, and each precedent for the defendant will block off an area where the factor definitely does not apply. This is shown in Figure~\ref{Trade}. We can now fit a line to the points and suggest that the factor applies to points  north east of the line and does not apply to points south west of the line. A possible line ($y = 120 - 10x$: designed to just include both precedents) is shown in Figure~\ref{Trade}. Of course, other lines are possible, and the function need not be linear, and so any unconstrained point may be the subject of debate as to whether or not the precedent applies.

The composite factor will appear as a node in the ADF with the dimensional facts as its children and the equation  will define the acceptance condition. This node can be treated as an issue for explanation purposes.

\section{Concluding Remarks}

In this paper we have discussed several reasons why cases are better seen as bundles of issues than as bundles of factors. The issues will be the bundles of factors. Thinking in terms of issues improves explanations by enabling them to focus on what was disputed and what is significant in the particular case under consideration, and to be expressed in the IRAC form widely taught in law schools. Using issues also greatly enhances factor based precedential constraint by eliminating irrelevant distinctions. The importance of issues for prediction was indicated by the central role given to issues in systems designed to predict legal decisions based  on factors such as \cite{bruninghaus2003predicting} ad \cite{grabmair2017predicting}, and is discussed in detail in \cite{bench2021}.

Developing good factor based explanations is of current importance because of the increased use of machine learning approaches for predicting legal decisions. Explanations are essential because of the right to explanation and to encourage acceptance of these predictions. But they also have relevance to the responsible use of such systems: machine learning approaches are vulnerable to changes in the law and social attitudes, and the bias that may exist in the past decisions. Good explanations will help to detect decisions based on reasons which are no longer applicable, and decisions based on reasons that exhibit bias. Explanation can therefore apply a corrective influence essential to the responsible use of Machine Learning in law.

%
%
%
%
\bibliographystyle{splncs04}
\bibliography{irac}

\end{document}